\documentclass[a4paper,conference]{IEEEtran}
\usepackage{cite}

\ifCLASSINFOpdf
\else
\fi
\usepackage{amsmath}
\usepackage{array}

\ifCLASSOPTIONcompsoc
  \usepackage[caption=false,font=normalsize,labelfont=sf,textfont=sf]{subfig}
\else
  \usepackage[caption=false,font=footnotesize]{subfig}
\fi
\usepackage{url}

\usepackage{todonotes}
\usepackage[export]{adjustbox}
\usepackage{amsmath}
\usepackage{makecell}
\begin{document}

\newcommand\Tstrut{\rule{0pt}{2.6ex}}  
\definecolor{mygreen}{rgb}{0.01, 0.75, 0.24}
\definecolor{green1}{rgb}{0.0, 0.5, 0.0}

%
\title{Improving Point Cloud Based Place Recognition with Ranking-based Loss and Large Batch Training}

\author{\IEEEauthorblockN{Jacek Komorowski}
\IEEEauthorblockA{Warsaw University of Technology\\
Warsaw, Poland\\
jacek.komorowski@pw.edu.pl}}


%


\maketitle

\begin{abstract}
The paper presents a simple and effective learning-based method for computing a discriminative 3D point cloud descriptor for place recognition purposes. 
Recent state-of-the-art methods have relatively complex architectures such as multi-scale pyramid of point Transformers combined with a pyramid of feature aggregation modules.
Our method uses a simple and efficient 3D convolutional feature extraction, based on a sparse voxelized representation, enhanced with channel attention blocks. 
We employ recent advances in image retrieval and propose a modified version of a loss function based on a differentiable average precision approximation. Such loss function requires training with very large batches for the best results. This is enabled by using multistaged backpropagation.
Experimental evaluation on the popular benchmarks proves the effectiveness of our approach, with a consistent improvement over the state of the art.
\end{abstract}


%
\IEEEpeerreviewmaketitle

\section{Introduction}
\label{sec:intro}

\begin{figure}
\centering
\includegraphics[width=1.0\linewidth,trim={0 6.9cm 14.5cm 0cm},clip]{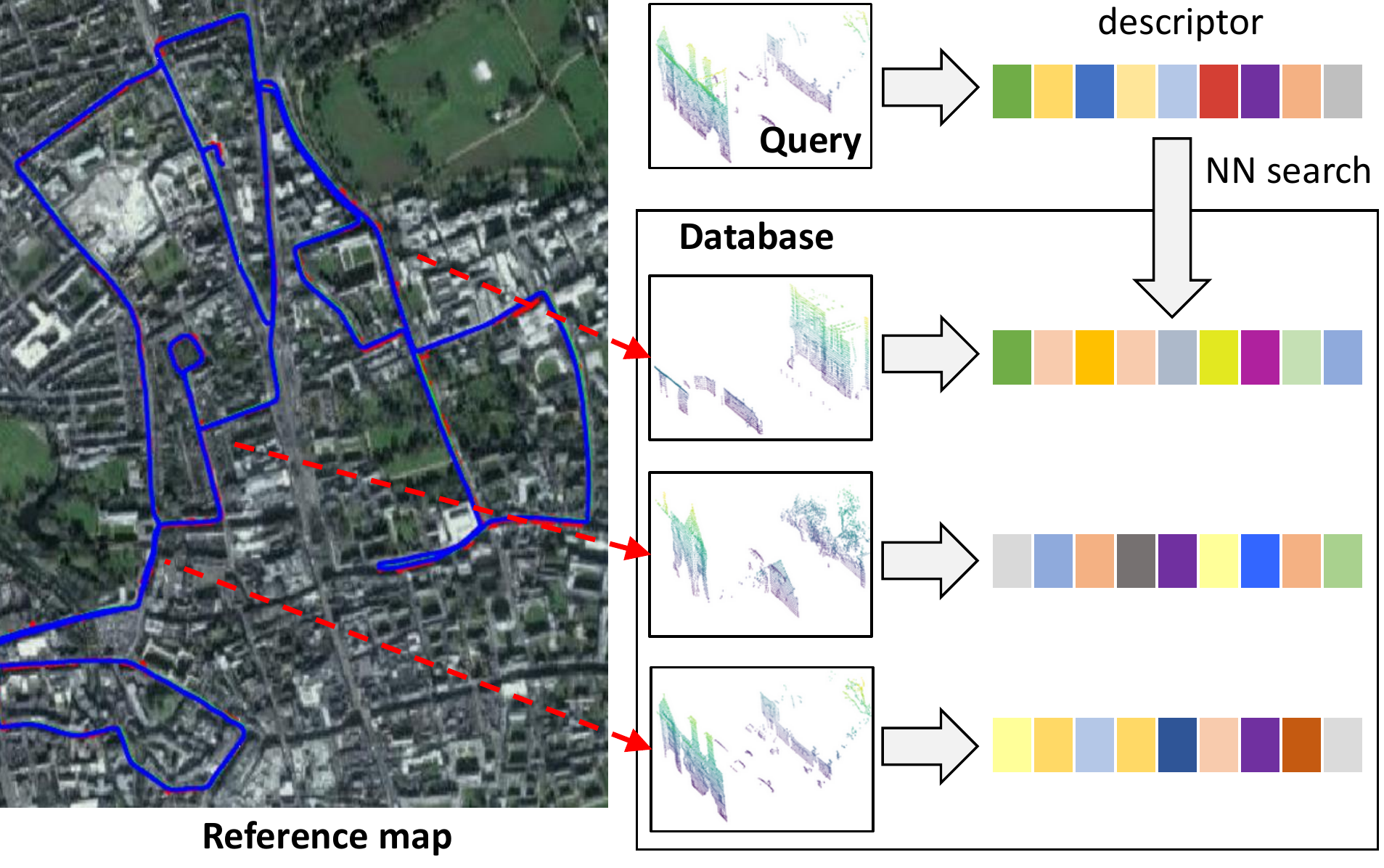}
\caption{Point cloud-based place recognition. The trained network computes a global descriptor of a query point cloud.
Localization is performed by searching the database for geo-tagged point clouds with closest descriptors.}
\label{fig:teaser}
\end{figure}

This work focuses on the point cloud-based place recognition using learned descriptors.
Geometry-based methods, using point clouds acquired by a LiDAR scanner, are gaining popularity as they are more robust to adverse environmental conditions compared to appearance-based solutions utilizing RGB images.
Place recognition can be formulated as an instance retrieval problem.
Given a query 3D point cloud, the goal is to retrieve its closest matches from a database. 
The localization of the query point cloud is reasoned based on a known coordinates of retrieved matches.
The search can be performed by using a trained network to compute a point cloud descriptor and finding point clouds with closest descriptors in the database, as shown in Fig.~\ref{fig:teaser}.

There's a long line of research on point cloud-based place recognition using learned descriptors~\cite{angelina2018pointnetvlad,zhang2019pcan,liu2019lpd,komorowski2020minkloc3d,9560932,hui2021pyramid,fan2021svt,xu2021transloc3d}, with a progressive advancements in state of the art.
PointNet-like architecture used in early works~\cite{angelina2018pointnetvlad,zhang2019pcan,liu2019lpd} was not well suited to extract informative features and was replaced by a 3D convolutional network based on a sparse voxelized representation in~\cite{komorowski2020minkloc3d}.
The contemporary methods~\cite{9560932,hui2021pyramid,fan2021svt,xu2021transloc3d} employ variants of Transformer~\cite{vaswani2017attention} architecture.
However, the progress has recently stagnated with little quantitative improvement between competitive methods. 
This can be partially explained by the limitation of the most popular benchmark introduced in PointNetVLAD~\cite{angelina2018pointnetvlad} paper, where top results are close to upper limits of the reported metrics.


However, the recent research has almost exclusively focused on the network architecture, resulting in relatively complex solutions combining pyramid Point Transformer with pyramid VLAD and Context Gating~\cite{hui2021pyramid}.
In this work we look on the problem from more holistic perspective and investigate the improvements in both the network architecture and the training process.
We show, that using an enhanced version of a simple 3D convolutional MinkLoc3D~\cite{komorowski2020minkloc3d} network, coupled with adoption of recent advances in ranking-based metric optimization allows outperforming contemporary methods with more complex architectures.
We adapted a loss function based on a smooth approximation of Average Precision metric~\cite{brown2020smooth} and combined it with multistage backpropagation\cite{revaud2019learning} to allow using very large training batches, consisting of thousands of point clouds.
Evaluation on the popular point cloud-based place recognition benchmark introduced in~\cite{angelina2018pointnetvlad} shows a consistent advantage over state of the art.
We release the code and trained models for the benefit of the community.
\footnote{Project web site: \url{https://github.com/jac99/MinkLoc3Dv2}}


\section{Related work}
\label{sec:related-work}

\subsection{Point cloud-based place recognition using learned descriptors}
PointNetVLAD \cite{angelina2018pointnetvlad} was the first neural network-based method to extract a global point cloud descriptor for large-scale place recognition purposes. 
It uses PointNet~\cite{qi2017pointnet} backbone to extract local features, followed by NetVLAD aggregation layer.
However, PointNet architecture is not well suited to extract informative local features that can be aggregated into a discriminative global descriptor.
LPD-Net~\cite{liu2019lpd} addressed this weakness by enhancing an input point cloud with handcrafted features and using graph neural networks to extract informative local features.
EPC-Net~\cite{hui2022efficient} improved upon LPD-Net by using proxy point convolutional neural network. 

Above methods representation a point cloud as an unordered set of 3D points.
MinkLoc3D~\cite{komorowski2020minkloc3d} proposed an efficient alternative based on the sparse voxelized representation.
It uses a simple 3D convolutional network with Feature Pyramid Network (FPN)~\cite{DBLP:journals/corr/LinDGHHB16} architecture to extract informative local features and aggregates them using GeM~\cite{radenovic2018fine} pooling into a discriminative global descriptor.
MinkLoc3D outperformed prior state of the art, while being computationally efficient.

Recent works employ Transformers-based attention mechanism~\cite{vaswani2017attention} to capture long-range dependencies and boost feature extraction effectiveness.
SVT-Net~\cite{fan2021svt} and TransLoc3D~\cite{xu2021transloc3d} add a Transformer module on top of the 3D sparse convolutional network to learn both short range local features and long-range contextual features.
NDT-Transformer~\cite{9560932} uses Transformers to extract features directly from a 3D Normal Distribution Transform (NDT) representation of the input point cloud. 
PPT-Net~\cite{hui2021pyramid} uses a pyramid of point transformers to extract local features capturing spatial relationships at different resolutions. Then, a pyramid VLAD module aggregates the multi-scale
features into a global descriptor.

In this work we show, that enhancing the simple convolutional architecture of MinkLoc3D and improving the training process, can produce state-of-the-art results and outperform all recent methods with more complex architectures.

\subsection{Loss function in the retrieval task}
Place recognition can be formulated as a retrieval task, where the goal is to order database elements by decreasing similarity to the query. Localisation of the query element is inferred from known coordinates of most similar database elements.
Ranking-based metrics used for the retrieval task evaluation, such as Average Recall (AR), are non-differentiable and cannot be directly optimized using the stochastic gradient descent.
Instead, surrogate loss functions that can be minimized using gradient descent are often employed. 
Popular choices are contrastive~\cite{bromley1994signature}, triplet~\cite{hermans2017defense} or quadruplet~\cite{chen2017beyond} losses.
A number of more sophisticated loss function was proposed in the following years, such as multi-similarity loss~\cite{wang2019multi} or proxy-anchor loss~\cite{kim2020proxy}.
However, experimental results in recent surveys~\cite{musgrave2020metric,roth2020revisiting} suggest that their advantage is limited dataset-dependent.
Indeed, prior works on point cloud-based place recognition use either lazy quadruplet loss  ~\cite{angelina2018pointnetvlad,zhang2019pcan,liu2019lpd,9560932,hui2021pyramid}
or triplet loss with batch-hard negative mining~\cite{komorowski2020minkloc3d,fan2021svt,xu2021transloc3d}.

The discrepancy between the optimization and evaluation metrics can leads to suboptimal performance of the trained model.
~\cite{brown2020smooth} presented a solution, a differentiable approximation of AveragePrecision (AP) metric by relaxing the non-differentiable Indicator function with a sigmoid function.
~\cite{patel2021recall} proposed a smooth approximation of Recall@k. 
To be effective, these differentiable approximations of ranking metrics must be computed on very large batches of training elements, with the size comparable to the evaluation set.
Such large batches, with thousands of high resolution images or point clouds, cannot be directly processed due to hardware constrains, such as limited GPU memory. 
To overcome this problem, \cite{revaud2019learning} proposed the \emph{multistaged backpropagation} algorithm. It allows calculating the loss function for very large batches by splitting the process into two parts.
First, gradient of the loss w.r.t. descriptors calculated separately for each batch element is computed.
Then, using the chain rule, the gradient of the loss w.r.t. the network weights is calculated.
This is done at the expense of increased training time, as each batch element needs to be processed twice by the network.
Inspired by these works, in this paper we propose a modified formulation of a differentiable Average Precision, called Truncated SoftAP, suitable for training place recognition descriptors.
Combined with multistaged backpropagation algorithm allowing training with very large batches, it leads to state-of-the-art results.

\section{Method description}
\label{sec:method}

This section presents our method to compute a discriminative point cloud descriptor for place recognition purposes. 
Place recognition is performed by searching the database of geo-tagged point clouds for descriptors closest, in the Euclidean distance sense, to the descriptor of the query element.  The idea is illustrated in Fig.~\ref{fig:teaser}.


\subsection{Network architecture}

The network architecture is an enhacement of the MinkLoc3D~\cite{komorowski2020minkloc3d} point cloud descriptor.
To improve the performance we increased the depth and width of the network, by adding additional convolutional and transposed convolution blocks and incremented the number of channels.
We also incorporated ECA~\cite{Wang_2020_CVPR} channel attention layers originally proposed for 2D convolutional networks.

\begin{figure}[!t]
\centerline{\includegraphics[scale=1.]{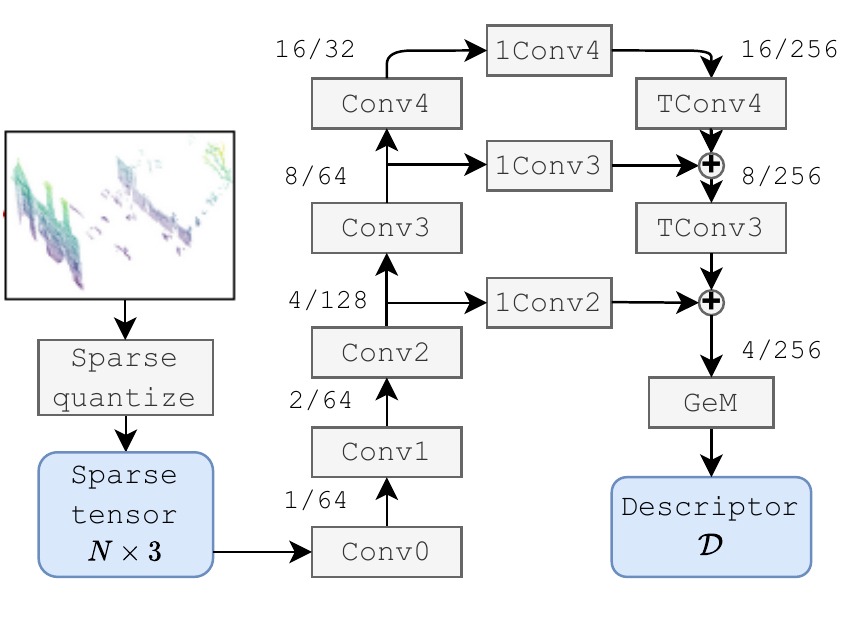}}
\caption{Overview of the descriptor extraction network. The input point cloud is quantized into a sparse 3D tensor
and processed by a 3D convolutional network with FPN architecture.
Conv are convolutional blocks, TConv are transposed convolutions and 1Conv are convolutions with 1x1x1 kernel.
The resultant feature map is aggregated using GeM pooling to produce a descriptor $\mathcal{D}$.
Numbers in parentheses (e.g. 2/64) show a stride and a number of channels of a feature map produced by each block.} 
\label{fig:high_level}
\end{figure}

The network has a sparse 3D convolutional architecture modelled on Feature Pyramid Network~\cite{lin2017feature}, shown in Fig.~\ref{fig:high_level}.
Using FPN architecture allows producing feature maps with increased receptive field, better capturing high-level semantic information, compared to the basic CNN design.
The input point cloud is first quantized into a single channel sparse tensor.
The bottom-up part consists of five convolutional blocks 
$\mathrm{Conv}_0, \ldots, \mathrm{Conv}_4$ 
producing sparse 3D feature maps with increasing receptive field.
Each convolutional block, starting from $\mathrm{Conv}_1$, decreases the spatial resolution by two. 
The top-down part consists of two transposed convolution layers $\mathrm{TConv}_3$ and $\mathrm{TConv}_4$.
Feature maps from higher pyramid levels, upsampled using transposed convolution, are added to the skipped features from the corresponding layer in the bottom-up trunk. 
Lateral connections use convolutions with 1x1x1 kernel (denoted $\mathrm{1Conv}$) to unify the number of channels produced by bottom-up blocks before they are merged in the top-down pass.
The resultant feature map is pooled using a generalized-mean (GeM) pooling~\cite{radenovic2018fine} layer to produce a descriptor $\mathcal{D}$. Details of each block are documented in Table~\ref{jk:tab-details}.
ECA is an Efficient Channel Attention~\cite{Wang_2020_CVPR} layer.
$\left[ \ldots \right]$ denotes a residual block. If an input number of channels to the residual block is different than an output number of channels, we add a convolution with 1x1x1 kernel to the residual connection.

\begin{table}[!t]
	\centering
	\caption{Details of the network architecture. BN=BatchNorm,
	ECA=Efficient Channel Attention layer, $\left[ \ldots \right]$ is a residual block.}
    \begin{tabular}{@{\enskip}l@{\quad}l@{\enskip}}
     \hline
    Block  &  Layers\\
     \hline
    \multicolumn{2}{c}{\textbf{Feature extraction}} \\
    {\tt Conv$_0$}  &  $c_0$ 5x5x5 filters - BN - ReLU\\
    {\tt Conv$_k$}, $k=1, 2, 3, 4$ &  $c_{k-1}$ 2x2x2  filters stride 2  - BN - ReLU \\
    &
    $\Big[$
    \makecell[l]{
       $c_k$ 3x3x3 filters stride 1 - BN - ReLU  \\
       $c_k$ 3x3x3 filters stride 1 - BN - ECA
    }
     $\Big]$
    \\
    & ReLU
    \\
    \multicolumn{2}{c}{where $c_0 = c_1=64, c_2=128, c_3=64, c_4=32$} \\
    {\tt TConv$_k$}, $k=3,4$ & 256 2x2x2 filters stride 2  \\
    {\tt 1Conv$_k$}, $k=2,3,4$ & 256 1x1x1 filters stride 1  \\
    \hline
    \multicolumn{2}{c}{\textbf{Feature aggregation}} \\
    {\tt GeM pooling} & Generalized-mean pooling layer \\
    \hline
	\end{tabular}
    \label{jk:tab-details}
\end{table}

\subsection{Network training}

\subsubsection{Loss function}

We train our network using a modified Smooth-AP loss~\cite{brown2020smooth}.
The original Smooth-AP aims at improving the ranking of all positives retrieved for each query element. 
In our context, \emph{positive} is a point cloud acquired at the same location as the query point cloud (max. 10\,m. distance) and a \emph{negative} is acquired at the different location (more than 50\,m. apart).
But for place recognition purposes, it's sufficient to retrieve only a small number of $k$ best candidates from the database.
It's not necessary to maximize the ranking of all positives. 
It's sufficient to ensure that top-$k$ candidates for each query are correct (positives).
We call our version of the loss function \emph{Truncated Smooth-AP} (TSAP), where the optimization goal is to maximize the average precision considering only top-$k$ positives.
Our version is more computationally efficient. The original Smooth-AP formulation requires computation of large $N \times P \times N$ similarity matrices, where $N$ is the batch size and $P$ the number of positives per query element.
In our approach, these matrices are $N \times k \times N$, where $k \ll P$.
This allows training with larger batch size, which further improves the accuracy.

Let $q$ be a query point cloud; $\mathcal{P}$ a set of its $k$ closest, in the descriptor space, positives; and $\Omega$ a set containing its all positives and negatives.
We use the following smooth average precision approximation:
\begin{equation}
    AP_q = \frac{1}{\left| P \right|} \sum_{i \in \mathcal{P}} 
    \frac
    {1 + \sum_{j \in \mathcal{P}, j \neq i} \mathcal{G} \left( d(q, i) - d(q, j) ; \tau \right) }
    {1 + \sum_{j \in \Omega, j \neq i} \mathcal{G} \left( d(q, i) - d(q, j) ; \tau \right) } \, ,
    \label{eq:softap}
\end{equation}
where $\mathcal{G}$ is a differentiable approximation of the indicator function, defined as 
$\mathcal{G} \left( x; \tau \right) = \left( 1 + \exp \left( -x/\tau \right) \right)^{-1}$;
and $d(q, i)$ is Euclidean distance between the descriptor of a query point cloud $q$ and $i$-th point cloud.
The parameter $\tau$ is the temperature controlling the sharpness of the approximation.
The nominator is a soft ranking of a positive $i$ among top-$k$ positives.
The denominator is a soft ranking of a positive $i$ among all other positives and negatives.

The loss function is computed for the batch of $m$ training elements. The batch is processed by the network to compute point clouds descriptors.
Each batch element is taken as a query $q$; a set of it's $k$ closest, in the descriptor space, positives ($\mathcal{P}$) and a set if its all positives and negatives ($\Omega$) are constructed by considering other elements in the batch.
The smooth average precision $AP_q$ within the batch with respect to $q$ is calculated.
The loss is an average for all batch elements defined as:
\begin{equation}
\mathcal{L}_{TSAP} = \frac{1}{m} \sum_{q=1}^{m} \left( 1 - AP_q \right) \, .
\end{equation}
The loss function is minimized using a stochastic gradient descent with Adam optimizer.

\subsubsection{Training with large batches}

Smooth approximations of a ranking-based metric, such as our Truncated Smooth-AP, are computed by considering the ranking within the batch of elements. 
For the optimal performance, they require large batch sizes comparable to the size of the evaluation set.
Processing large batches with thousands of of high dimensional data items is not feasible due to a limited GPU memory. 
To overcome this problem, we use the \emph{multistaged backpropagation} approach, proposed in~\cite{revaud2019learning}. The loss calculation is split into multiple stages.
First, each batch element is processed separately to compute its descriptor, with gradient computation switched off.  Then, the loss function is computed using descriptors of all elements in the forward pass and the gradient of the loss w.r.t. the descriptors is calculated in the backward pass.
Finally, a descriptor of each batch element is recalculated with gradient computation switched on.
The chain rule is used to combine results of earlier computations to get the gradient of the loss w.r.t. network weights. 
Gradients are accumulated, one batch element at a time, before finally updating the network weights.
For more detailed description of multistaged backpropagation see~\cite{revaud2019learning}.

\subsubsection{Implementation details}
We quantize 3D point coordinates using 0.01 quantization step. As point coordinates are normalized in $\left[-1, 1\right]$ range, this gives 200 voxel resolution in each spatial coordinate.
Parameters of the training process are listed in~Tab.\ref{jk:tab:params}.
Initial learning rate is divided by 10 at the epochs given by LR scheduler steps parameter. 
The dimensionality of the resultant global descriptor is set to 256, same as in other evaluated methods.

\begin{table}[htbp]
\caption{Training parameters in Baseline and Refined protocols.}
\begin{center}
\begin{tabular}{l@{\quad}l@{\quad}l}
\hline
& \begin{tabular}{@{}c@{}}Baseline \end{tabular}
& \begin{tabular}{@{}c@{}}Refined\end{tabular}
\\
[2pt]
\hline
Batch size & 2048 & 2048 \\
Number of epochs & 400 &  500 \\
Initial learning rate & 1e-3 & 1e-3 \\
LR scheduler steps & 250, 350 &  350, 450 \\
$L_2$ weight decay & 1e-4 &  1e-4 \\
Sigmoid temperature $(\tau)$ & 0.01 & 0.01  \\
Positives per query $(k)$ & 4 & 4 \\
[2pt]
\hline
\end{tabular}
\end{center}
\label{jk:tab:params}
\end{table}

To reduce overfitting, we apply data augmentation. 
It includes random jitter with a value drawn from $\mathcal{N}\left(0;0.001 \right)$; random translation by a value sampled from $\mathcal{U}\left(0, 0.01 \right)$; and removal of up to 10\% of randomly chosen points.

All experiments are performed on a server with a single nVidia RTX 2080Ti GPU, 12 core AMD Ryzen Threadripper 1920X processor and 64 GB of RAM. 
We use PyTorch 1.10~\cite{NEURIPS2019_9015} deep learning framework, MinkowskiEngine 0.5.4~\cite{choy20194d} auto-differentiation library for sparse tensors and Pytorch Metric Learning library 1.1~\cite{musgrave2020metric}.

\section{Experimental results}

This section describes results of an experimental evaluation of our point cloud descriptor and comparison with state of the art.


\subsection{Datasets and evaluation methodology}

Our method is trained and evaluated using datasets and evaluation protocols introduced in~\cite{angelina2018pointnetvlad}.  
This is a standard benchmark used in recent works on the point cloud-based place recognition in the outdoor environment:~\cite{angelina2018pointnetvlad,zhang2019pcan,liu2019lpd,sun2020dagc,komorowski2020minkloc3d,2016particular,fan2021svt,9560932,xu2021transloc3d,hui2021pyramid,hui2022efficient}.
It contains point clouds extracted from Oxford RobotCar dataset~\cite{RobotCarDatasetIJRR} and three in-house datasets: University Sector (U.S.), Residential Area (R.A.), Business District (B.D.). 
Point clouds are acquired using a LiDAR sensor mounted on the car travelling at different times of day and year through the same routes at these four areas. 
Oxford dataset is built by concatenating consecutive 2D scans from SICK LMS-151 2D LiDAR during the 20 m. drive of a vehicle.
In-house datasets contain point clouds acquired using Velodyne HDL-64 3D LiDAR. 
Each dataset is divided into disjoint training and test splits. 
Point clouds are preprocessed by removing the ground plane and downsampling to 4096 points. 
The point coordinates are rescaled and shifted to be within the $\left[-1, 1\right]$ range. 
An exemplary preprocessed point clouds are shown in Fig.~\ref{fig:search_results}. 

We follow two evaluation protocols introduced in~\cite{angelina2018pointnetvlad}.
In \emph{baseline} protocol, the network is trained using the training split of Oxford RobotCar dataset and evaluated on test splits of Oxford and in-house datasets.
In \emph{refined} protocol, the network is trained using a larger and more diverse set consisting of training splits of Oxford and in-house datasets.
The number of training and test elements in each scenario is shown in Tab.~\ref{jk:tab:datasets}.

\begin{table}[htbp]
\caption{Number of elements in training and test splits in Baseline and Refined evaluation protocols.}
\begin{center}
\begin{tabular}{l@{\quad}r@{\quad}r@{\quad}r@{\quad}r}
\hline
& \multicolumn{2}{c}{Baseline protocol} 
& \multicolumn{2}{c}{Refined protocol} 
\\
& \begin{tabular}{@{}c@{}}Training \end{tabular}
& \begin{tabular}{@{}c@{}}Test \end{tabular}
& \begin{tabular}{@{}c@{}}Training \end{tabular}
& \begin{tabular}{@{}c@{}}Test \end{tabular}
\\
[2pt]
\hline
Oxford & 21.7k & 3.0k & 21.7k & 3.0k \\
In-house
& \begin{tabular}{@{}c@{}} - \end{tabular} & 4.5k & 6.7k & 1.7k \\
[2pt]
\hline
\end{tabular}
\end{center}
\label{jk:tab:datasets}
\end{table}

Two point clouds are considered structurally similar for training purposes, if they are at most 10m apart based on the ground truth UTM coordinates. 
Point clouds are structurally dissimilar, if they are at least 50m apart. 
During evaluation, a point cloud from a test split is taken as a query and point clouds from different traversals that cover the same region form the database. 
The query point cloud is successfully localized if at least one of the top $N$ retrieved database clouds is within 25 meters from the ground truth position of the query.
\emph{Recall@N} is defined as the percentage of correctly localized queries.
In addition to Average Recall@1 (AR@1) we report Average Recall@1\% (AR@1\%), which is calculated taking into account top-$k$ matches, where $k$ is 1\% of the database size.
For detailed description of datasets and evaluation protocols please refer to~\cite{angelina2018pointnetvlad}.

\subsection{Results and discussion}
\label{sec:results}

\begin{figure*}[!t]
\centering
\setcounter{subfigure}{0}
\subfloat[\textbf{query}]{\includegraphics[width=2.8cm,height=2.1cm,trim={1.5cm 1.6cm 0.5cm 2.2cm},clip,cfbox=black 0.5pt 0.5pt]{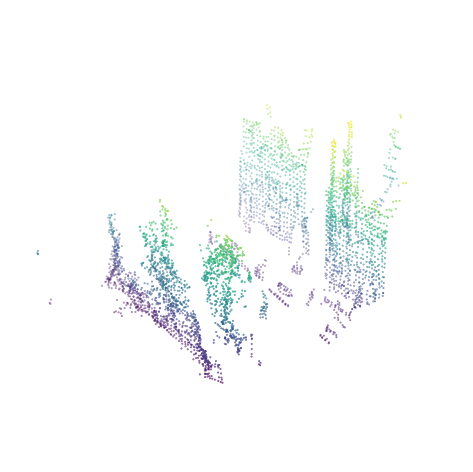}}
\hfill
\subfloat[dist=1.10 TP]{\includegraphics[width=2.8cm,height=2.1cm,trim={1.5cm 1.6cm 0.5cm 2.2cm},clip,cfbox=mygreen 0.5pt 0.5pt]{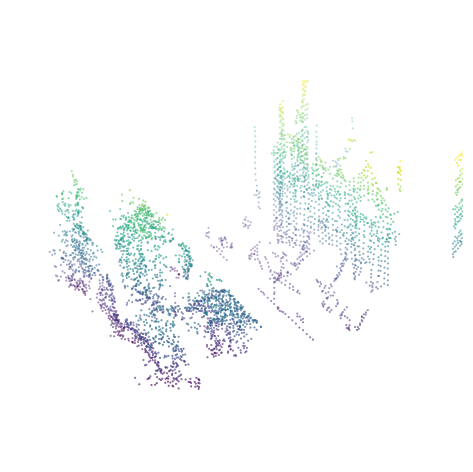}}
\hfill
\subfloat[dist=1.19 TP]{\includegraphics[width=2.8cm,height=2.1cm,trim={1.5cm 1.6cm 0.5cm 2.2cm},clip,cfbox=mygreen 0.5pt 0.5pt]{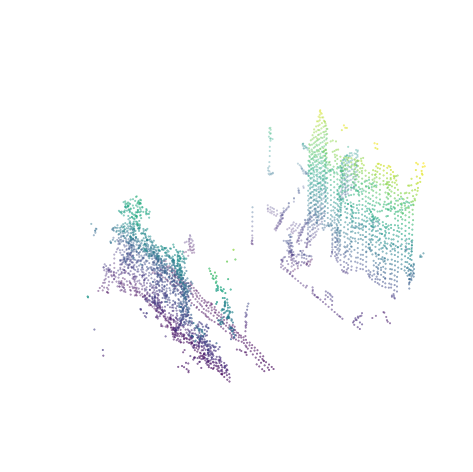}}
\hfill
\subfloat[dist=1.36 TP]{\includegraphics[width=2.8cm,height=2.1cm,trim={1.5cm 1.6cm 0.5cm 2.2cm},clip,cfbox=mygreen 0.5pt 0.5pt]{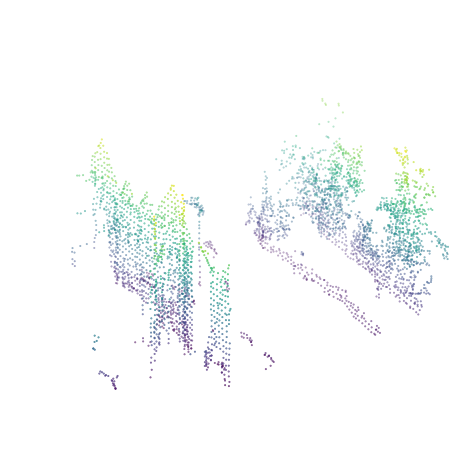}}
\hfill
\subfloat[dist=1.53 TP]{\includegraphics[width=2.8cm,height=2.1cm,trim={1.5cm 1.6cm 0.5cm 2.2cm},clip,cfbox=mygreen 0.5pt 0.5pt]{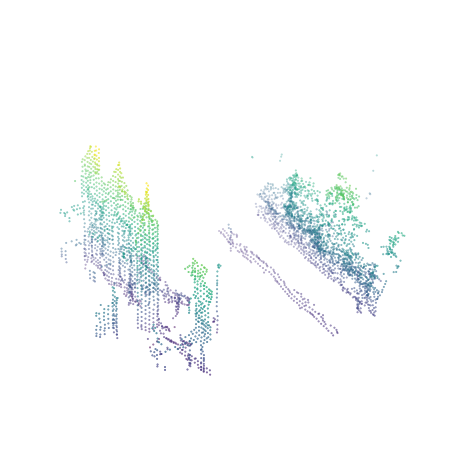}}
\hfill
\subfloat[dist=1.56 FP]{\includegraphics[width=2.8cm,height=2.1cm,trim={1.5cm 1.6cm 0.5cm 2.2cm},clip,cfbox=red 0.5pt 0.5pt]{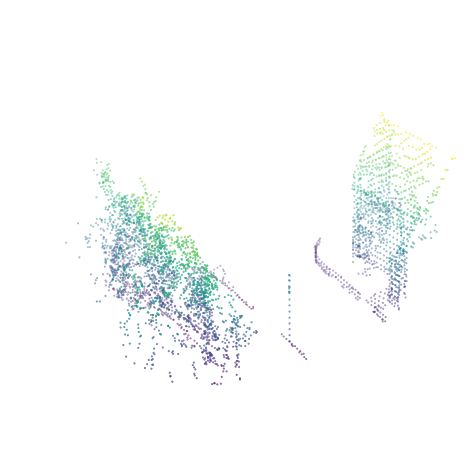}}
\\[-1ex]
\subfloat[\textbf{query}]{\includegraphics[width=2.8cm,height=2.1cm,trim={1.5cm 1.6cm 0.5cm 2.2cm},clip,cfbox=black 0.5pt 0.5pt]{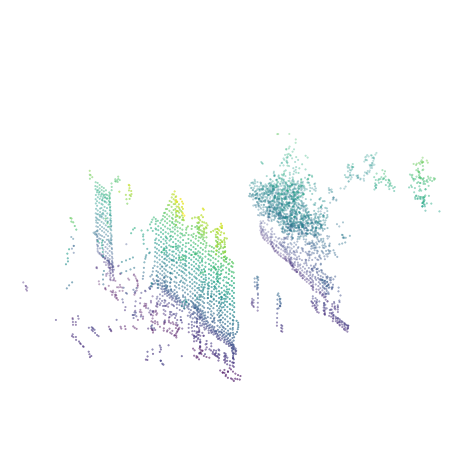}}
\hfill
\subfloat[dist=0.89 TP]{\includegraphics[width=2.8cm,height=2.1cm,trim={1.5cm 1.6cm 0.5cm 2.2cm},clip,cfbox=mygreen 0.5pt 0.5pt]{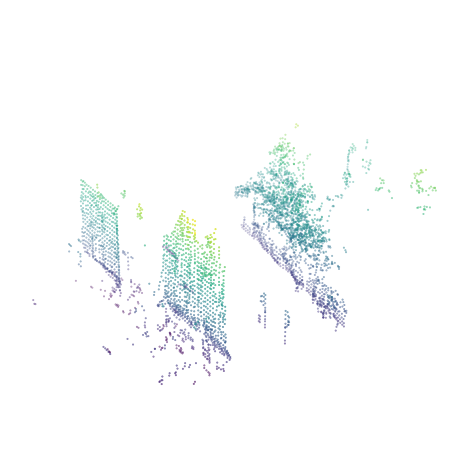}}
\hfill
\subfloat[dist=1.48 TP]{\includegraphics[width=2.8cm,height=2.1cm,trim={1.5cm 1.6cm 0.5cm 2.2cm},clip,cfbox=mygreen 0.5pt 0.5pt]{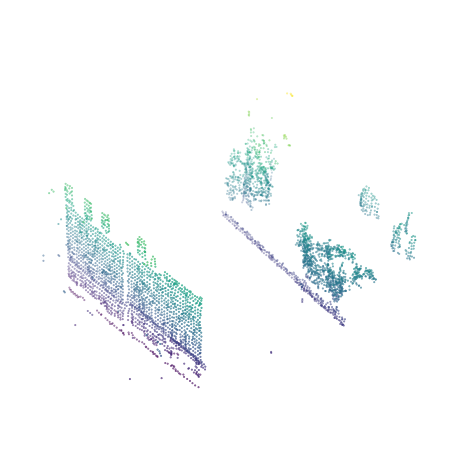}}
\hfill
\subfloat[dist=1.50 FP]{\includegraphics[width=2.8cm,height=2.1cm,trim={1.5cm 1.6cm 0.5cm 2.2cm},clip,cfbox=red 0.5pt 0.5pt]{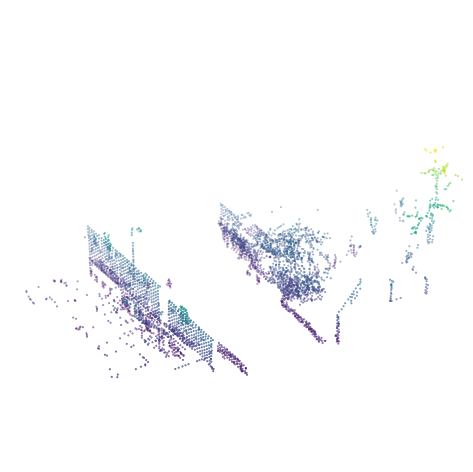}}
\hfill
\subfloat[dist=1.52 FP]{\includegraphics[width=2.8cm,height=2.1cm,trim={1.5cm 1.6cm 0.5cm 2.2cm},clip,cfbox=red 0.5pt 0.5pt]{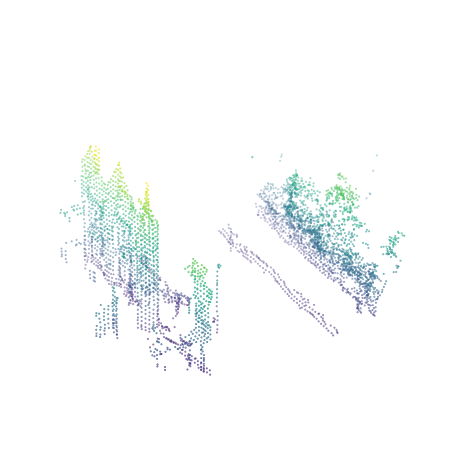}}
\hfill
\subfloat[dist=1.57 FP]{\includegraphics[width=2.8cm,height=2.1cm,trim={1.5cm 1.6cm 0.5cm 2.2cm},clip,cfbox=red 0.5pt 0.5pt]{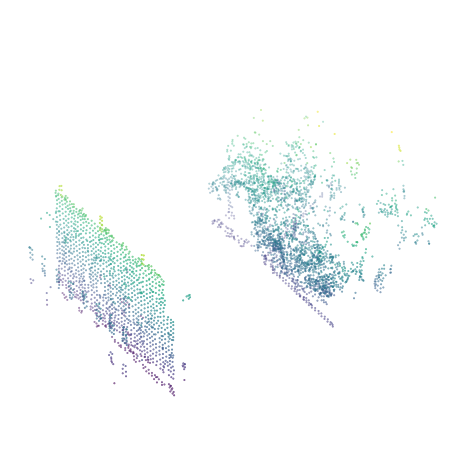}}
\caption{Nearest neighbour search results. The leftmost column shows a query point cloud. Other columns show its five nearest neighbours. \emph{dist} is an Euclidean distance in the descriptor space. TP indicates true positive and FP false positive.}
\label{fig:search_results}
\end{figure*}

\begin{figure}[!t]
\centering
\subfloat{\includegraphics[width=2.8cm,height=2.2cm,trim={1.5cm 1.6cm 0.5cm 2.2cm},clip,cfbox=black 0.5pt 0.5pt]{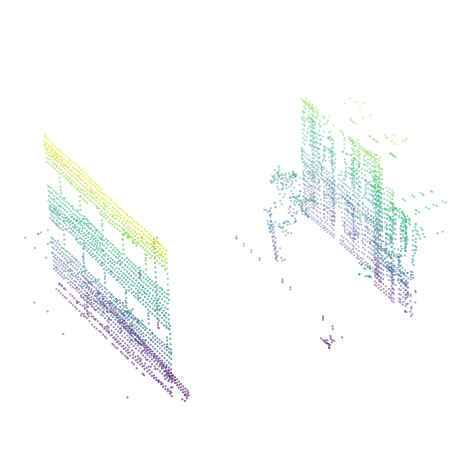}}
\hfill
\subfloat{\includegraphics[width=2.8cm,height=2.2cm,trim={1.5cm 1.6cm 0.5cm 2.2cm},clip,cfbox=red 0.5pt 0.5pt]{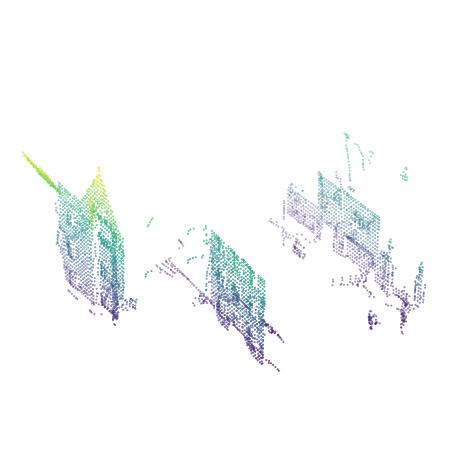}}
\hfill
\subfloat{%
\includegraphics[width=2.8cm,height=2.2cm,trim={1.5cm 1.6cm 0.5cm 2.2cm},clip,cfbox=mygreen 0.5pt 0.5pt]{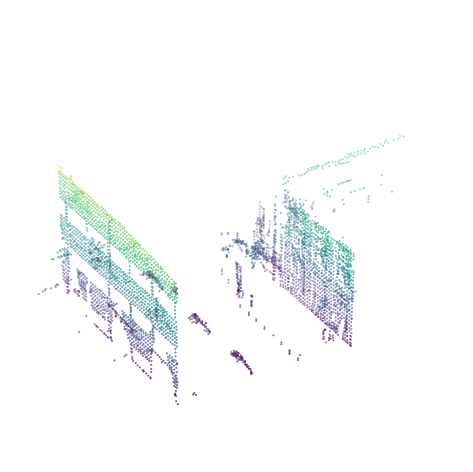}}
\\[-1ex]
\setcounter{subfigure}{0}
\subfloat[(a) query]{\includegraphics[width=2.8cm,height=2.2cm,trim={1.5cm 1.6cm 2cm 2.2cm},clip,cfbox=black 0.5pt 0.5pt]{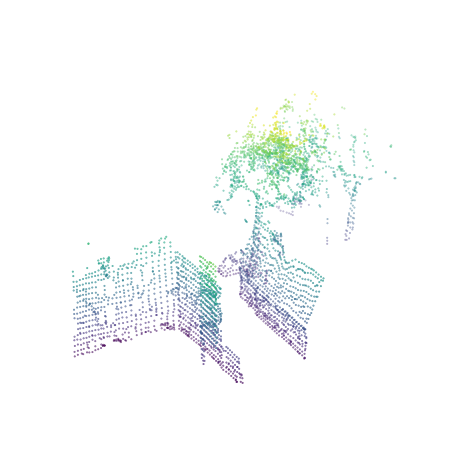}}
\hfill
\subfloat[(b) Incorrect match]{\includegraphics[width=2.8cm,height=2.2cm,trim={1.5cm 1.6cm 2cm 2.2cm},clip,cfbox=red 0.5pt 0.5pt]{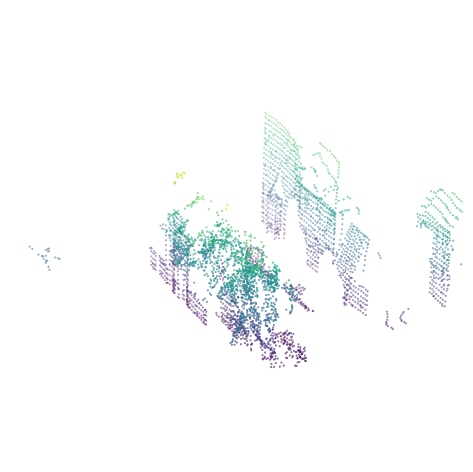}}
\hfill
\subfloat[(c) Closest positive]{\includegraphics[width=2.8cm,height=2.2cm,trim={1.5cm 1.6cm 2cm 2.2cm},clip,cfbox=mygreen 0.5pt 0.5pt]{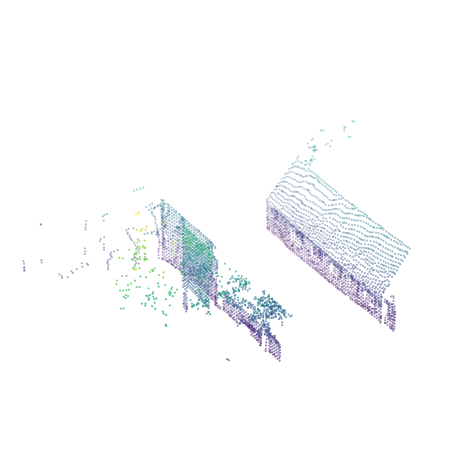}}
\caption{Failure cases. Examples of unsuccessful retrieval results using our method. (a) is the query point cloud, (b) incorrect match to the query and (c) the closest true match. }
\label{fig:failure_cases}
\end{figure}

Figure~\ref{fig:search_results} visualizes nearest neighbour search results using our descriptor in Oxford evaluation subset. The leftmost column shows a query point cloud and other columns show its five nearest neighbours. 
Figure~\ref{fig:failure_cases} shows failure cases. 

\captionsetup[subfigure]{labelformat=empty}

\subsubsection{Comparison with state of the art.}
Our methods yields a consistent improvement over state of the art on almost all datasets and evaluation metrics.
For Baseline evaluation protocol (Table~\ref{jk:tab:results_baseline}), we improve Average Recall@1 metric on Oxford dataset by 1.3\,p.p. to 96.3\% and on three in-house datasets (U.S., R.A., B.D.) by 0.8 to 2.2\,p.p.
It must be noted that in the Baseline protocol the method is trained using data from a training split of Oxford dataset only. It generalizes to in-house datasets better than competitive methods, although a performance drop can be noticed. This can be attributed to the fact, that in-house datasets were acquired using a LiDAR with a different characteristics.

Bigger improvement is achieved for Refined evaluation protocol (Table~\ref{jk:tab:results_refine}). Average Recall@1 metric on Oxford dataset is 96.9\% and outperforms runner-up (TransLoc3D) by 1.9\,p.p. On three in-house datasets (U.S., R.A., B.D.) we outperform TransLoc3D by 1.0 to 2.8\,p.p.

Results for AR@1\% metric for Refined evaluation protocol are almost saturated and there's little room for improvement. Nevertheless, our method improves the results on Oxford by 0.6\,p.p. to reach 99.1\% and on Business District (B.D.) by 2.7\, p.p. to reach 99.1\%.
AR@1\% averaged over all 4 datasets improves by 0.4\% to 99.3\% compared to the runner-up (TransLoc3D).

\captionsetup[subfigure]{labelformat=parens}

\begin{table*}[htbp]
\caption{Evaluation results (Average Recall at 1\%) of place recognition methods trained using Baseline protocol. Numbers in color show the difference with the best competitive method.}
\begin{center}
\begin{tabular}{l@{\quad}l@{\quad}l@{\quad}l@{\quad}l@{\quad}l@{\quad}l@{\quad}l@{\quad}l|@{\quad}l@{\quad}l}
\hline
& \multicolumn{2}{c}{Oxford} & \multicolumn{2}{c}{U.S.}  & \multicolumn{2}{c}{R.A.} & \multicolumn{2}{c}{B.D.} & \multicolumn{2}{c}{Mean} \\
& \begin{tabular}{@{}c@{}}AR@1 \end{tabular}
& \begin{tabular}{@{}c@{}}AR@1\% \end{tabular}
& \begin{tabular}{@{}c@{}}AR@1 \end{tabular}
& \begin{tabular}{@{}c@{}}AR@1\% \end{tabular}
& \begin{tabular}{@{}c@{}}AR@1 \end{tabular}
& \begin{tabular}{@{}c@{}}AR@1\% \end{tabular}
& \begin{tabular}{@{}c@{}}AR@1 \end{tabular}
& \begin{tabular}{@{}c@{}}AR@1\% \end{tabular}
& \begin{tabular}{@{}c@{}}AR@1 \end{tabular}
& \begin{tabular}{@{}c@{}}AR@1\% \end{tabular}
\\
[2pt]
\hline
PointNetVLAD~\cite{angelina2018pointnetvlad} &  62.8 & 80.3 & 63.2 & 72.6 & 56.1 & 60.3 & 57.2 & 65.3 & 59.8 & 69.6  \\
PCAN~\cite{zhang2019pcan} & 69.1 & 83.8 & 62.4 & 79.1 & 56.9 & 71.2 & 58.1 & 66.8 & 61.6 & 75.2 \\
LPD-Net~\cite{liu2019lpd} & 86.3 & 94.9 & 87.0 & 96.0 & 83.1 & 90.5 & 82.5 & 89.1 & 84.7 & 92.6 \\
EPC-Net~\cite{hui2022efficient} & 86.2 & 94.7 & -  & 96.5 & - & 88.6 & - & 84.9 & - & 91.2 \\
SOE-Net~\cite{xia2020soe} & - & 96.4 & - & 93.2 & - & 91.5 & - & 88.5 & - & 92.4 \\
MinkLoc3D~\cite{komorowski2020minkloc3d} & 93.0 & 97.9 & 86.7 & 95.0 & 80.4 & 91.2 & 81.5 & 88.5 & 85.4 & 93.2 \\
NDT-Transformer~\cite{9560932} & 93.8 & 97.7 & - & - & - & - & - & - & - & - \\
PPT-Net~\cite{hui2021pyramid} & 93.5 & 98.1 & 90.1 & \textbf{97.5} & 84.1 & 93.3 & 84.6 & 90.0 & 88.1 & 94.7 \\
SVT-Net~\cite{fan2021svt} & 93.7 & 97.8 & 90.1 & 96.5 & 84.3 & 92.7 & 85.5 & 90.7 & 88.4 & 94.4 \\
TransLoc3D~\cite{xu2021transloc3d} & 95.0 & 98.5 &  - & 94.9 & - & 91.5 & - & 88.4 & - & 93.3 \\
MinkLoc3Dv2 (ours) & \textbf{96.3} {\color{green1} +1.3} & \textbf{98.9} {\color{green1} +0.4} & \textbf{90.9} {\color{green1} +0.8} & 96.7 {\color{red} -0.8} & \textbf{86.5} {\color{green1} +2.2} & \textbf{93.8} {\color{green1} +0.5} & \textbf{86.3} {\color{green1} +0.8} & \textbf{91.2} {\color{green1} +0.5} & \textbf{90.0} {\color{green1} +1.6} & \textbf{95.1} {\color{green1} +0.4} \\
[2pt]
\hline
\end{tabular}
\end{center}
\label{jk:tab:results_baseline}
\end{table*}


\begin{table*}[htbp]
\caption{Evaluation results (Average Recall at 1\% and at 1) of place recognition methods trained using Refined protocol. Numbers in color show the difference with the best competitive method.}
\begin{center}
\begin{tabular}{l@{\quad}l@{\quad}l@{\quad}l@{\quad}l@{\quad}l@{\quad}l@{\quad}l@{\quad}l|@{\quad}l@{\quad}l}
\hline
& \multicolumn{2}{c}{Oxford} & \multicolumn{2}{c}{U.S.}  & \multicolumn{2}{c}{R.A.} & \multicolumn{2}{c}{B.D.} & \multicolumn{2}{c}{Mean}\\
& \begin{tabular}{@{}c@{}}AR@1 \end{tabular}
& \begin{tabular}{@{}c@{}}AR@1\% \end{tabular}
& \begin{tabular}{@{}c@{}}AR@1 \end{tabular}
& \begin{tabular}{@{}c@{}}AR@1\% \end{tabular}
& \begin{tabular}{@{}c@{}}AR@1 \end{tabular}
& \begin{tabular}{@{}c@{}}AR@1\% \end{tabular}
& \begin{tabular}{@{}c@{}}AR@1 \end{tabular}
& \begin{tabular}{@{}c@{}}AR@1\% \end{tabular}
& \begin{tabular}{@{}c@{}}AR@1 \end{tabular}
& \begin{tabular}{@{}c@{}}AR@1\% \end{tabular}
\\[2pt]
\hline
PointNetVLAD~\cite{angelina2018pointnetvlad} & 63.3 & 80.1 & 86.1 & 94.5 & 82.7 & 93.1 & 80.1 & 86.5 & 78.0 & 88.6 \\
PCAN~\cite{zhang2019pcan} & 70.7 & 86.4 & 83.7 & 94.1 & 82.5 & 92.5 & 80.3 & 87.0 & 79.3 & 90.0 \\
DAGC~\cite{sun2020dagc} & 71.5 & 87.8 & 86.3 & 94.3 & 82.8 & 93.4 & 81.3 & 88.5 & 80.5 & 91.0 \\
LPD-Net~\cite{liu2019lpd} & 86.6 & 94.9 & 94.4 & 98.9 & 90.8 & 96.4 & 90.8 & 94.4 & 90.7 & 96.2 \\
SOE-Net~\cite{xia2020soe} & 89.3 & 96.4 & 91.8 & 97.7 & 90.2 & 95.9 & 89.0 & 92.6 & 90.1 & 95.7 \\
MinkLoc3D~\cite{komorowski2020minkloc3d} & 94.8 & 98.5 & 97.2 & 99.7 & 96.7 & 99.3 & 94.0 & 96.7 & 95.7 & 98.6 \\
PPT-Net~\cite{hui2021pyramid} & - & 98.4 & - & 99.7 & - & 99.5 & - & 95.3 & - & 98.2 \\
SVT-Net~\cite{fan2021svt} & 94.7 & 98.4 & 97.0 & \textbf{99.9} & 95.2 & 99.5 & 94.4 & 97.2 & 95.3 & 98.8 \\
TransLoc3D~\cite{xu2021transloc3d} & 95.0 & 98.5 & 97.5 & 99.8 & 97.3 & \textbf{99.7} & 94.8 & 97.4 & 96.2 & 98.9 \\
MinkLoc3Dv2 (ours) & \textbf{96.9} {\color{green1} +1.9} & \textbf{99.1} {\color{green1} +0.6} & \textbf{99.0} {\color{green1} +1.5} & 99.7 {\color{red} -0.2} & \textbf{98.3} {\color{green1} +1.0} & 99.4 {\color{red} -0.3} & \textbf{97.6} {\color{green1} +2.8} & \textbf{99.1} {\color{green1} +1.7} & \textbf{97.9\textbf} {\color{green1} +1.7} & \textbf{99.3}  {\color{green1} +0.4} \\
[2pt]
\hline
\end{tabular}
\end{center}
\label{jk:tab:results_refine}
\end{table*}



\subsubsection{Ablation study}
In this section we investigate impact of the enhancements in the network architecture and the training procedure improvements on the performance of our point cloud descriptor.
In these experiments the network is trained using the Refined protocol and evaluated on Oxford and three in-house datasets (U.S., R.A. and B.D.).
As the baseline we use MinkLoc3D~\cite{komorowski2020minkloc3d} architecture trained with a triplet loss and batch-hard negative mining~\cite{hermans2017defense}.
The results are summarized in Table~\ref{jk:tab:ablation}.
Improving the network architecture, without changing the loss function, improves the performance, measured as a mean Average Recall@1 (AR@1) over all evaluation sets, by 2.1\,p.p. to 96.6\%.
Using our TruncatedSoftAP loss and large batch size with multistage backpropagation, increases AR@1 of the baseline architecture by 2.5\,p.p. to 97.0\%.
Using both improved architecture and training procedure leads to overall 3.4\,p.p. gain over the baseline (97.9\% AR@1).
It can be noticed, that TruncatedSoftAP loss produces the best results when used with large training batches (2048 elements). When trained with smaller batches (256 elements), AR@1 is worse by app. 1\,p.p.



\begin{table}[htbp]
\begin{center}
\begin{tabular}{c@{\;}l@{\quad}l@{\quad}r@{\quad}r@{\quad}r}
\hline
& \begin{tabular}{@{}c@{}}Architecture\end{tabular}
& \begin{tabular}{@{}c@{}}Loss function\end{tabular}
& \begin{tabular}{@{}c@{}}Batch \\size\end{tabular}
& \begin{tabular}{@{}c@{}}Mean\\AR@1\end{tabular}
& \begin{tabular}{@{}c@{}}Mean\\AR@\%\end{tabular}
\\
[2pt]
\hline
& MinkLoc3D & Triplet & 256 & 94.5 & 98.4 \\
& MinkLoc3D & TruncatedSmoothAP & 256 & 96.0 & 98.8 \\
& MinkLoc3D & TruncatedSmoothAP & 2048 & 97.0 & 99.0 \\
&  MinkLoc3Dv2 & Triplet & 256 &  96.6 & 98.9 \\
& MinkLoc3Dv2 & TruncatedSmoothAP & 256 &  96.7 & 99.1 \\
* & MinkLoc3Dv2 & TruncatedSmoothAP & 2048 & \textbf{97.9} & \textbf{99.3} \\
[2pt]
\hline
\end{tabular}
\end{center}
\caption{Impact of a network architecture and the training procedure on the performance of the point clouds descriptor. The network is trained using Refined protocol. * indicates our final method.}
\label{jk:tab:ablation}
\end{table}

\section{Conclusion}
\label{sec:conclusions}

In this paper we present a method for computing a discriminative point cloud descriptor for place recognition purposes. Our method uses a relatively simple 3D convolutional architecture with improved training procedure. We propose a loss function based on a modified version of a smooth average precision metric combined with very large training batches. Experimental evaluation proves that our method consistently outperforms state of the art.

That achieved results (AR@1 between 96.9\% and 99.4\% for Refined evaluation protocol) show that standard benchmarks are saturated and there's very little room for improvement.
New datasets, covering much larger areas and more diverse environments, are needed to promote and benchmark further progress in the field.

\bibliographystyle{IEEEtran}
\bibliography{my-bib}

\end{document}